%% file: main.tex
\documentclass[lettersize,journal]{IEEEtran}

\input{header}

\title{Using Models Based on Cognitive Theory to Predict Human Behavior in Traffic: A Case Study}

\author{Julian F. Schumann, Aravinda Ramakrishnan Srinivasan, Jens Kober, Gustav Markkula, Arkady Zgonnikov
\thanks{The authors are with the Department of Cognitive Robotics, Delft University of Technology, Delft, Zuid Holland 2628 CD, The Netherlands (e-mail: j.f.schumann@tudelft.nl; j.kober@tudelft.nl; a.zgonnikov@tudelft.nl) and the Institute for Transport Studies, Leeds University, Leeds, The United Kingdom (e-mail: A.R.Srinivasan@leeds.ac.uk; g.markkula@leeds.ac.uk) \textit{(Corresponding Author: Julian Schumann)}}
\thanks{This research was partially supported by TAILOR, a project funded by EU Horizon 2020 research and innovation programme under GA No 952215 and by the UK EPSRC grant EP/S005056/1. For the purpose of open access, the author(s) has applied a Creative Commons Attribution (CC BY) license to any Accepted Manuscript version arising.}
\thanks{The source code, trained models, and data can be found online at a \href{https://github.com/julianschumann/Commotions-model-evaluation}{public Github repository}. This includes \href{https://github.com/julianschumann/Commotions-model-evaluation/blob/main/Framework/Supplementary_material.pdf}{supplementary materials}.}}
\usepackage{layouts}

\begin{document}
\twocolumn
\maketitle
\begin{abstract}
    The development of automated vehicles has the potential to revolutionize transportation, but they are currently unable to ensure a safe and time-efficient driving style. Reliable models predicting human behavior are essential for overcoming this issue. While data-driven models are commonly used to this end, they can be vulnerable in safety-critical edge cases. This has led to an interest in models incorporating cognitive theory, but as such models are commonly developed for explanatory purposes, this approach's effectiveness in behavior prediction has remained largely untested so far. In this article, we investigate the usefulness of the \emph{Commotions} model -- a novel cognitively plausible model incorporating the latest theories of human perception, decision-making, and motor control -- for predicting human behavior in gap acceptance scenarios, which entail many important traffic interactions such as lane changes and intersections. We show that this model can compete with or even outperform well-established data-driven prediction models across several naturalistic datasets. These results demonstrate the promise of incorporating cognitive theory in behavior prediction models for automated vehicles.
\end{abstract}

\section{Introduction}
Automated vehicles have become a major focus of the car industry in recent years due to their potential to revolutionize transportation. The promised benefits of automated vehicles include fewer accidents caused by human errors, increased accessibility of mobility solutions, and more efficient use of time while traveling~\cite{brar_impact_2017, meyer_autonomous_2017, pisarov_future_2021}. However, despite significant investments~\cite{holland-letz_mobilitys_2021}, there are still only prototypes of automated vehicles on the street, and they are not yet widely available to the public~\cite{milford_self-driving_2020, wang_safety_2020}. One major challenge to the widespread adoption of automated vehicles is ensuring that they are both efficient and safe, traveling in a timely and efficient manner while also maintaining a level of safety that is at least equivalent to human driving~\cite{milford_self-driving_2020, sinha_crash_2021}. However, many automated vehicles currently focus on ensuring safety, avoiding any action that could potentially lead to an accident. While this approach may reduce the risk of traffic participants being harmed, it misses out on travel efficiency and acceptance, requiring further efforts to make automated vehicles truly useful~\cite{milford_self-driving_2020, wang_safety_2020}. One potential solution is to incorporate prediction models to reduce uncertainty about future human behavior and allow for more actions to be classified as safe~\cite{sadigh_planning_2016, mozaffari_deep_2022}. 

Accurate predictions of human behavior are especially critical in scenarios involving gap acceptance~\cite{schumann_benchmarking_2023}, which form a significant subset of space-sharing conflicts in traffic, including situations such as crossing an intersection or changing lanes~\cite{markkula_defining_2020}. Many models for predicting human behavior in these scenarios have been developed, including trajectory prediction models~\cite{salzmann_trajectron_2020, yuan_agentformer_2021, zhang_forceformer_2023} and models predicting the binary choice of either accepting or rejecting the gap~\cite{pekkanen_variable-drift_2022, theofilatos_cross_2021,xie_data-driven_2019}. However, most of these include few assumptions about human decision-making -- using a mainly data driven approach known for being unreliable in safety-critical edge cases~\cite{marcus_next_2020, schumann_benchmarking_2023}. 

Meanwhile, there is a separate literature of cognitive theory developed to explain human decision-making in traffic~\cite {markkula_explaining_2022, zgonnikov_should_2022}. Inclusion of such theory into predictive models might help overcome the unreliability issues of purely data-driven approaches~\cite{marcus_next_2020}. 
However, current cognitively plausible models have a number of limitations which hinder their use for behavior prediction. In particular, most such models are limited to a specific scenario~\cite{pekkanen_variable-drift_2022,zgonnikov_should_2022} and cannot handle complex inputs which prevents their applications to naturalistic datasets. As a result, it is currently unknown if incorporating cognitive theories in behavior prediction models could actually yield any benefits in terms of prediction accuracy and robustness.

This study aims to explore the potential of one possible approach of incorporating cognitive theory into prediction models: the adaption of a specific existing explanatory model~\cite{markkula_explaining_2022} to function as a prediction model, using gap acceptance as target scenario type. Adaptation of this model for prediction purposes is non-trivial, and does in itself represent a significant contribution to the field (Section~\ref{sec:Commotions_adaptations}). Furthermore, we also conduct an ablation study to find the most promising configurations for the model (Section~\ref{sec:exp_I}). Finally. we compare the performance of the resulting configurations of this model to state-of-the-art data-driven prediction models (Section~\ref{sec:exp_III}). 

\section{Background}
This section provides a description of the general type of gap acceptance scenarios addressed here, a brief overview of the tested cognitive model, and an introduction to a framework that facilitates unbiased comparisons of the model's predictive performance against existing benchmarks. We also discuss our changes of the tested model that enable its use as a predictive model.

\subsection{Gap acceptance} \label{sec:Gap acceptance}
Gap acceptance problems are a type of traffic interaction that involves a space-sharing conflict between two agents with intersecting paths, such as intersections, pedestrian crossings, and lane changes on highways~\cite{schumann_benchmarking_2023, markkula_defining_2020}. There, these two agents can be differentiated by the possession of the right of way, with the vehicle with priority being referred to as the ego vehicle $V_E$. In such a situation, the other agent, designated as the target vehicle $V_T$, must then decide whether to cross $V_E$'s path in front of $V_E$ (i.e., accepting the offered gap) or to wait until $V_E$ has passed, thereby rejecting the gap. For example, if $V_T$ approaches an intersection via a secondary road, it needs to decide whether the gap to the vehicle coming from the perpendicular direction is large enough to cross the intersection without waiting for that car to pass (Fig.~\ref{fig:commotions}). Accurately predicting $V_T$'s decision in such scenarios is crucial for $V_E$, as $V_T$'s future behavior could limit $V_E$'s options, such as $V_E$ being forced to slow down to prevent a collision by $V_T$ accepting the gap.

\subsection{The \emph{Commotions} model} \label{sec:Commotions}
\begin{figure}
\centering
\includegraphics[width = \linewidth]{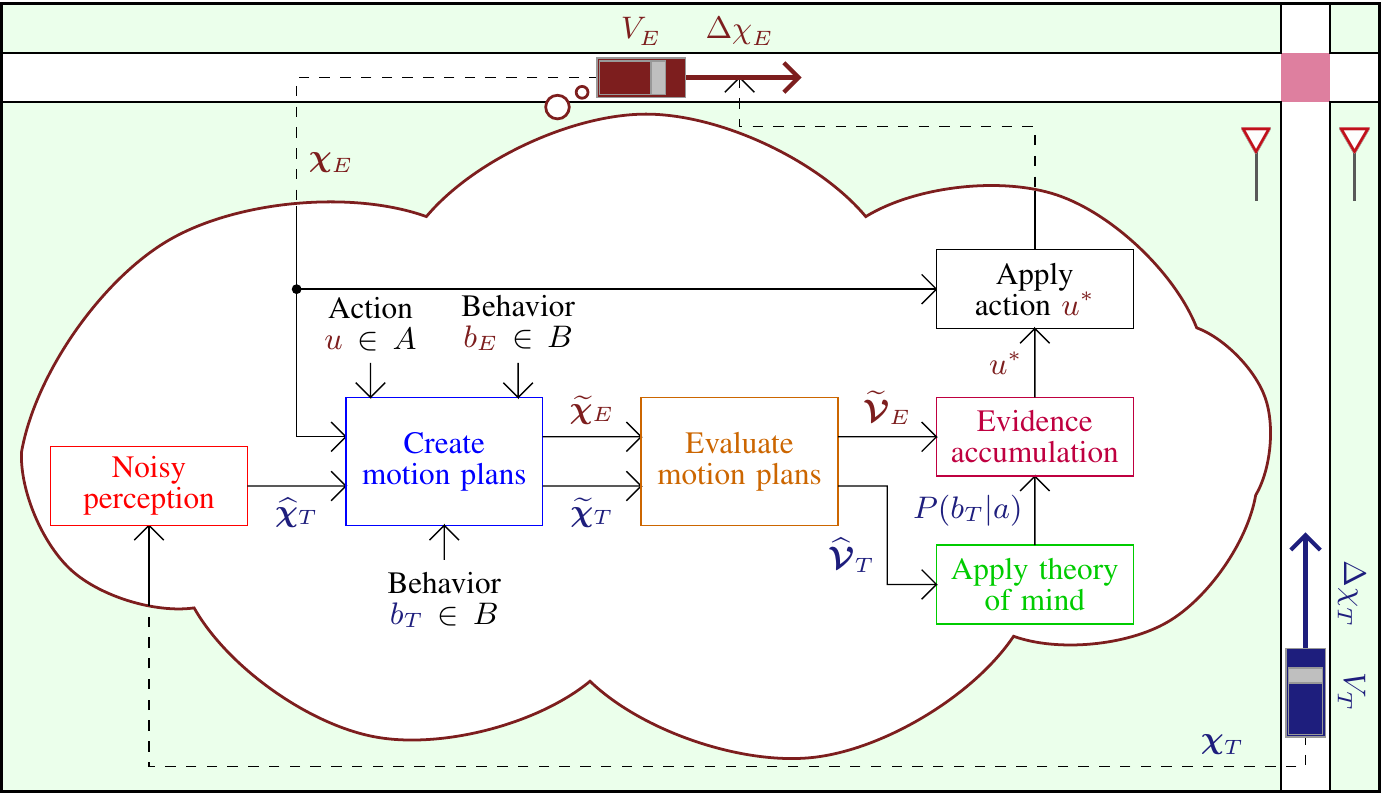}
\vspace{-5mm}
\caption{A depiction of the \emph{Commotions} model~\cite{markkula_explaining_2022} and its high-level parts, showing how the position of ego vehicle $V_E$ is updated at one point in time. Simultaneously, the target vehicle $V_T$ also updates its kinematic state by using the same mechanics -- only with mirrored inputs.}
\label{fig:commotions}
\vspace{-5mm}
\end{figure}

Markkula et al.~\cite{markkula_explaining_2022} proposed a cognitive framework for modeling road user interactions in gap acceptance scenarios. Their framework includes a wide range of cognitive mechanisms, such as decision-making based on evidence accumulation~\cite{pekkanen_variable-drift_2022}, noisy perception~\cite{kwon_unifying_2015} and applying a theory of mind~\cite{chen_drivers_2019} (Fig.~\ref{fig:commotions}). They implemented this framework in models for interactions between vehicles and/or pedestrians on straight crossing paths, i.e., including gap acceptance scenarios between two vehicles. What we will refer to here as the \emph{Commotions} model (after the name of the project in which the model was developed) is the most successful model variant identified in \cite{markkula_explaining_2022}, applied to such scenarios.

As illustrated in Fig.~\ref{fig:commotions}, the proposed model postulates that at each time step, both ego vehicle $V_E$ and target vehicle $V_T$ concurrently determine their current control inputs. This decision-making process of each agent is subject to sensory noise and Bayesian filtering during the perception of the position of the other agent. Based on their own short-term control input $u$ ($A$ is a discrete set) and both vehicles' long-term behavior $b_E$ and $b_T$ (i.e., preference for going first or second through the contested space), corresponding pairs of future trajectories -- represented by pairs of $\widetilde{\bm{\chi}}_E$ and $\widetilde{\bm{\chi}}_T$ -- are generated, with the constraint that the resulting interactions are safe. Each pair of trajectories is then evaluated (punishing large control inputs, time delays, and traffic rule violations), resulting in the value 
$\widetilde{\mathcal{V}}_E$ representing the agent's own opinion and the value $\widehat{\mathcal{V}}_T$, which is the value the agent assumes that the other agent assigns to each trajectory pair for each possible combination of behaviors and control inputs. Each agent then weighs the evaluation $\widetilde{\mathcal{V}}_E$ of their own trajectory based on the probability of the other vehicle behaving accordingly, assuming per the theory of mind that this probability is correlated with the respective value $\widehat{\mathcal{V}}_T$. Evidence accumulation is used to ensure no abrupt and seemingly arbitrary changes in behavior, by combining the weighted values with previous evaluations of a potential action $u$ and only changing the applied control input $u^*$ if this entails a sufficiently substantial improvement in this accumulated value, i.e., control is intermittent. Based on the currently chosen control input $u^*$ each agent's states are then projected forward to the next time step. By repeatedly using this process for both agents, the model can generate a pair of simulated trajectories on the perpendicular intersection. To represent the models randomness, $n_p$ different trajectory pairs are generated in repeated simulations.

\subsection{The framework for benchmarking gap acceptance models} \label{sec:Framework}
To compare several prediction models in a fair and unbiased manner, we utilize a framework previously developed by Schumann et al.~\cite{schumann_benchmarking_2023}. This framework facilitates the comparison of such models in any gap acceptance scenario according to a wide selection of metrics. Moreover, it grants precise control over the timing of the evaluated predictions and the allocation of individual samples to training and testing sets.

The framework also permits the conversion of different types of predictions, including between binary and trajectory predictions, increasing the number of metrics that can be employed to compare models. For instance, the benchmark enables models that originally predict only gap acceptance probabilities to also generate predictions of full trajectories. Specifically, to transform a predicted probability $a_{\text{pred}} \in [0,1]$ of accepting the gap into a set of predicted trajectories for a given sample, the framework uses two instances of a state-of-the-art trajectory prediction model~\cite{salzmann_trajectron_2022}. One of these models is trained exclusively on samples with accepted gaps, while the other is trained on samples with rejected gaps. Both models are utilized to predict a set of trajectories based on the given sample's input, from which the final set is sampled with weights adjusted by $a_{\text{pred}}$~\cite{schumann_benchmarking_2023}.

\section{Commotions as a predictive model} \label{sec:Commotions_adaptations}
Although the \emph{Commotions} model's capability of expressing a number of empirically observed human interaction phenomena was demonstrated successfully in the original paper~\cite{markkula_explaining_2022}, it was not developed for use as a prediction model. As such, it has many limitations compared to existing models developed for this purpose. For once, the computational efficiency of its existing implementation makes training and testing on most datasets infeasible. In this paper, we address this problem by implementing parallel processing of multiple model predictions on a GPU and using analytical instead of numerical integration inside the model. This achieves a speed increase of roughly four orders of magnitude. 

\label{sec:commotions_prediction}
Another problem with the \emph{Commotions} model is that it is constrained to the scenario of perpendicular intersections with straight trajectories seen in Fig.~\ref{fig:commotions}, which is incongruent to most real world situations. We utilize an expansion of the benchmarking framework~\ref{sec:Framework} allowing us to project real-world two-dimensional trajectories onto the quasi-one-dimensional-scenario required as the input data. Namely, for each agent, we define a method for determining the most probable path from their current location towards the contested space, where the trajectories of the ego vehicle $V_E$ and the target vehicle $V_T$ intersect. The length of this path is then assumed to be equal to the distances of those agents to the contested space (the purple square in Fig.~\ref{fig:commotions}) along the respective perpendicular streets. 

While it might be possible to use the same approach to project predicted trajectories from the quasi-one-dimensional scenario to the original two-dimensional space, they would only be projected onto the aforementioned predefined most probable paths. As this would drastically limit the solution space, we instead use scenario-independent information from the predicted trajectories. For each pair of trajectories from simulation $p$ we can determine if the contested space was reached first by $V_T$ ($a_{\text{pred},p} = 1$ represents an accepted gap) or $V_E$ ($a_{\text{pred},p} = 0$). Simultaneously, the time $t_{A, \text{pred}, p}$ of $V_T$ reaching the contested space can be extracted as well. Averaging over all predictions allows us to calculate the probability $a_{\text{pred}} \in[0,1]$ of $V_T$ accepting the gap. Combined with the predicted time of acceptance $t_A$, which the framework accepts as another type of prediction~\cite{schumann_benchmarking_2023}, generating predicted trajectories in the original space then becomes possible (\ref{sec:Framework}).

Finally, the \emph{Commotions} model is able to process merely the current position and velocity of only the two principal actors in a gap acceptance scenario, i.e., $V_E$ and $V_T$, and not any other agents in the scene. However, as this only hinders but does not prevent the model's predictive usage, this issue remains currently unaddressed.

\section{Evaluating configurations of the \\ Commotions model} \label{sec:exp_I} \label{sec:commotions_comp}
In this section, we investigate the predictive performance of several configurations of the \emph{Commotions} model stemming from a number of design decisions that have to be made when using the \emph{commotions} model to predict human behavior. For example, the modeling of the interaction between $V_E$ and $V_T$ can utilize either an \emph{interactive} approach, where both agents utilize all aspects of the \emph{Commotions} model (Fig.~\ref{fig:commotions}) to determine their current control inputs $u^*$ (\emph{IM}), or a \emph{non-interactive} approach (\emph{NM}), where only the behavior of $V_T$ is predicted by the model, with $V_E$ set to maintaining its original velocity. Meanwhile, another decision pertains to selecting the form of short-term control inputs $u$, with the options being the application of either a \emph{constant acceleration} (\emph{AC}) or \emph{constant jerk} (\emph{JC}).

As important parts of the model such as the creation of the trajectories $\bm{\chi}_E$ and $\bm{\chi}_T$ (Fig.~\ref{fig:commotions} and \ref{sec:Commotions}) are non-differentiable, we use Bayesian optimization~\cite{jones_efficient_1998} to fit the \emph{Commotions} model's parameters. However, regarding the optimization procedure, some open questions still remain. First, the user must decide whether to train the model in a \emph{single optimization} round (\emph{1O}) or use a \emph{two-stage optimization} (\emph{2O}), wherein the second stage of optimization is carried out over a reduced parameter search space surrounding the optimized parameters obtained in the first stage. Second, a choice between the two available \emph{loss functions} $\mathcal{L}_1$ and $\mathcal{L}_2$ used to fit the \emph{Commotions} model's parameters must be made. $\mathcal{L}_1$ is adapted directly from the work of Zgonnikov et al.\cite{zgonnikov_should_2022} (with $t_C$ being the time when $V_E$ reaches the purple intersection in Fig.~\ref{fig:commotions}) and evaluates every prediction $p$ for each sample $i$, while $\mathcal{L}_2$ expands upon this by enforcing more varied predictions:
\begin{equation}\begin{aligned}
    \mathcal{L}_1 = & \sum\limits_{i} \frac{1}{n_p} \sum\limits_{p = 1}^{n_p} 4 \left\vert a_i - a_{\text{pred},i,p} \right\vert + \left(t_{i} - t_{A,\text{pred},i,p} \right)^2\\
    \mathcal{L}_2 = & \,\mathcal{L}_1 + \sum\limits_{i} 100 V_{i} - 20 \sqrt{V_{i}} + 1 \\
    & {\Eqonesep} = \min\left\{t_{A,i}, \max\left\{t_{C,i},  t_{A,\text{pred},i,p}\right\} \right\} \\
    & V_i = \min \left\{ \mathbb{V}_p\left(t_{A,\text{pred},i,p}\right), \frac{1}{100}\right\}
\end{aligned}\end{equation}
 
\subsection{Setup} \label{sec:setup_I}
\begin{table*}
    \caption{Assessing the influence of the binary configuration choices in the \emph{Commotions} model (\ref{sec:commotions_comp}) on the predictive performance.\parnote[]{The individual results underlying the values shown here can be found in the form of figures and tables in the \href{https://github.com/julianschumann/Commotions-model-evaluation/blob/main/Framework/Supplementary_material.pdf}{supplementary materials}. In the two right-most columns, statistical significance of the differences in metrics is tested with a paired Student t-test (significance level $\alpha = 0.05$). The first number in each cell represents percentage of cases on the randomly split testing sets, whereas the number in the parentheses corresponds to the critical split (i.e., the testing set including the most unintuitive samples). In the last row, results are split by metric.}} 
    \begin{tabularx}{\textwidth}{C|DEF|G|H|H}
    \toprule
    Binary choice $\mathcal{C}$ & $\mathcal{C}_1$ & vs & $\mathcal{C}_2$ & \multicolumn{1}{c|}{Cases} & $\mathcal{C}_1$ better than  $\mathcal{C}_2$ & $\mathcal{C}_2$ better than  $\mathcal{C}_1$ \\ \midrule
    Ego vehicle's modeling & Interactive (\emph{IM}) & vs & Non-interactive (\emph{NM}) & 88 & $\Ho 7 \%$ ($7 \%$)$\Ho$ & $\Ho 2 \%$ ($22 \%$)\\ \midrule
    Vehicles' control input & Jerk (\emph{JC}) & vs & Acceleration (\emph{AC}) & 88 & $20 \%$ ($23 \%$) & $26 \%$ ($25 \%$)\\ \midrule
    Optimization method & Single round (\emph{1O}) & vs & Two rounds (\emph{2O}) & 88 & $\Ho 2 \%$ ($15 \%$) & $\Ho 6 \%$ ($11 \%$)\\ \midrule
    \multirow{2}{*}{Optimized loss function} & \multirow{2}{*}{$\mathcal{L}_1$} & \multirow{2}{*}{vs} & \multirow{2}{*}{$\mathcal{L}_2$} & 24 (\emph{ADE}) & $79 \%$ ($29 \%$) & $\hphantom{0}0 \%$ ($13 \%$) \\
    &&&& 64 (Other) & $\hphantom{0}0  \%$ ($11 \%$) & $52 \%$ ($30 \%$)\\\bottomrule
    \end{tabularx} 
    \parnotes
    \label{tab:commmotions_comp} 
    \vspace{-8mm}
\end{table*}
\subsubsection{Datasets}
The predictive performance of the different model configurations is compared using three datasets, each focusing on a different scenario.
\begin{itemize}
\item \emph{L-GAP}~\cite{zgonnikov_should_2022}, a driving simulator dataset, contains scenarios in which $V_T$ must decide whether to turn left in front of or behind $V_E$ approaching on the opposite lane.
\item \emph{rounD}~\cite{krajewski_round_2020}, a real-world dataset captured by a drone, covers roundabouts where $V_T$ must decide whether to enter the roundabout in front of or behind $V_E$ which is already inside the roundabout.
\item The \emph{UDISS} dataset~\cite{srinivasan_commotions_2023}, created in a driving simulator, focuses on a perpendicular intersection where $V_T$ must cross either in front of or behind $V_E$, which is driving along the other road with the right of way.
\end{itemize}
While the latter two datasets include other agents besides $V_E$ and $V_T$, in this paper we ignore those due to the aforementioned limitations of the \emph{Commotions model}, with the resulting datasets being referred to respectively as $\text{\emph{rounD}}_{\text{2V}}$ and $\text{\emph{UDISS}}_{\text{2V}}$. We also restrict the provided input trajectories to two input time steps ($n_I = 2$), as this provides sufficient information to extract the two agents' current positions and velocities, which are the only inputs the \emph{Commotions} model is able to process.

\subsubsection{Train/test splits}
On each dataset, we perform eleven training-and-testing cycles for each configuration. In ten of these, the split between training and testing set is random. In the last split however, we place the samples that exhibit the most unintuitive human behavior -- smallest accepted gaps and largest rejected gaps -- into the \textit{critical} testing set. This latter approach allows us to evaluate the robustness of the model's predictive capabilities against the most challenging and safety-critical cases.

\subsubsection{Metrics} \label{sec:setup_I_metrics}
To evaluate the models' predictions made on the testing set, we employ three metrics which have previously been used to assess different aspects of gap acceptance predictions~\cite{schumann_benchmarking_2023, theofilatos_cross_2021, salzmann_trajectron_2020}. First, the area under the receiver-operator curve (\emph{AUC}) assesses binary predictions (accept/reject gap) at two different time points: the initial opening of the gaps and the time corresponding to a fixed (dataset-specific) characteristic gap size~\cite{schumann_benchmarking_2023}. Second, the average displacement error (\emph{ADE}) metric evaluates full predicted trajectories at the characteristic gap size. Third, we use the true negative rate under perfect recall~\cite{schumann_benchmarking_2023} (\emph{TNR-PR}), a metric that rates the usefulness of binary predictions made on the smallest possible gaps at the last point in time when they can aid in adjusting $V_E$'s planned path accordingly. 
However, due to a lack of gaps accepted after this point in time on the \emph{UDISS} dataset, the \emph{TNR-PR} cannot be calculated on that scenario, resulting in eleven viable combinations of metrics and datasets we can use to compare model configurations. 

Furthermore, when we transform binary predictions into trajectory predictions, so that for example the \emph{ADE} metric can be applied to the \emph{Commotions} model, we use \emph{Trajectron++}~\cite{salzmann_trajectron_2020}, a state-of-the-art trajectory prediction model, in accordance to the method laid out in Section~\ref{sec:Framework}.

\subsection{Results} \label{sec:results_I}
Following the setup described above, we test 16 configurations of the \emph{Commotions} model (resulting from four independent design choices) on the eleven combinations of datasets and metrics, resulting in 88 comparisons for each design choice on both random and critical split test sets. For example, on the \emph{L-GAP} dataset, the \emph{AUC} averaged over the ten random test sets for predictions made at the fixed-size gap (a size of \SI{3.36}{s}) ranges from \num{0.936} to the value \num{0.970} produced by $\text{\emph{CM}}_{NA12}$, which utilizes the non-interactive modeling approach (\emph{NM}) and acceleration control (\emph{AC}) and was trained in one round of optimizing (\emph{1O}) $\mathcal{L}_2$.

Comparison between the configurations of the Commotions model (Tab.~\ref{tab:commmotions_comp}) indicates that there was no consistently better alternative for any of the four design choices. Still, we are able to make some recommendations. For example, the non-interactive modeling approach (\emph{NM}) appears to be more likely to outperform its opposite on the critical test set, while having the added advantage of faster evaluations by obviating half of the \emph{Commotions} model's calculations updating $\bm{\chi}_E$~(Fig.\ref{fig:commotions}). Similarly, using acceleration control (\emph{AC}) produces better predictions slightly more often, possibly by enabling the model to predict faster human reactions. Although the number of optimization rounds appears to be largely irrelevant, using only one round of optimization (\emph{1O}) may make the model even more robust on the critical test sets, with faster training being another benefit. Comparatively, the most significant factor seems to be the choice of the loss function -- as long as one differentiates by metric. Specifically, $\mathcal{L}_1$ is a better choice when minimizing \emph{ADE}, while $\mathcal{L}_2$ is superior on the other three metrics. This is expected, as the regularization achieved by $\mathcal{L}_2$ enforcing some variance in the predictions also leads to a larger spread of predicted trajectories, resulting in a larger average displacement error.

When seeking the best configuration of the \emph{Commotions} model, rather than comparing the binary choices, we can compare the 16 configurations among themselves as well, either by the average result over the ten random test sets or the result on the critical test sets. As model performance mainly depends on the chosen metric, here we discuss \emph{ADE} separately from other metrics. Specifically, we found that the $\text{\emph{CM}}_{NA11}$ configuration is best, having a lower \emph{ADE} in $79\%$ of all the 90 possible comparisons -- i.e, on two types of results, three datasets, and against 15 other configurations. Using the same approach on the remaining metrics, we find the most promising configuration to be $\text{\emph{CM}}_{NA12}$ with better metric values in $70\%$ of all cases. These results further support $\text{\emph{CM}}_{NA11}$ and $\text{\emph{CM}}_{NA12}$ (non-interactive modeling, acceleration vehicle input, single-round optimization) as the optimal configurations of the \emph{Commotions} model.

\section{Comparing Commotions to Established Models}  \label{sec:exp_III}
In this section, we assess the potential of the \textit{Commotions} model by comparing the predictive performance of two of its configurations ($\text{\emph{CM}}_{NA11}$ for the \emph{ADE} and $\text{\emph{CM}}_{NA12}$ for other metrics) against established prediction models. Besides the \emph{Trajectron++} model (\emph{T++}) introduced in Section~\ref{sec:setup_I_metrics}, we also used a logistic regression model (\emph{LR}) as a baseline, with both methods having previously demonstrated good performance on similar gap acceptance problems~\cite{schumann_benchmarking_2023}. While these models have far fewer restrictions on the type of input data they can process, we artificially constrain the used input data to the \emph{Commotions} model's limitations to allow for an equitable comparison. The only exception is the dimensionality of the input for \emph{T++}, as this model can only process the original two-dimensional trajectories, but not the projected quasi-one-dimensional inputs of the \emph{Commotions} model. 

\subsection{Setup}  \label{sec:setup_III}
Regarding the chosen datasets, testing and training splits as well as the chosen metric, this experiment follows the setup of the previous ablation study (\ref{sec:setup_I}). Within the same setup, we evaluated the two configurations of the \emph{Commotions} model ($\text{\emph{CM}}_{NA11}$ and $\text{\emph{CM}}_{NA12}$) against the state-of-the-art models (\emph{T++} and two versions of \emph{LR}).

\subsection{Results}  \label{sec:results_III}
\begin{figure}
    \centering
    \includegraphics{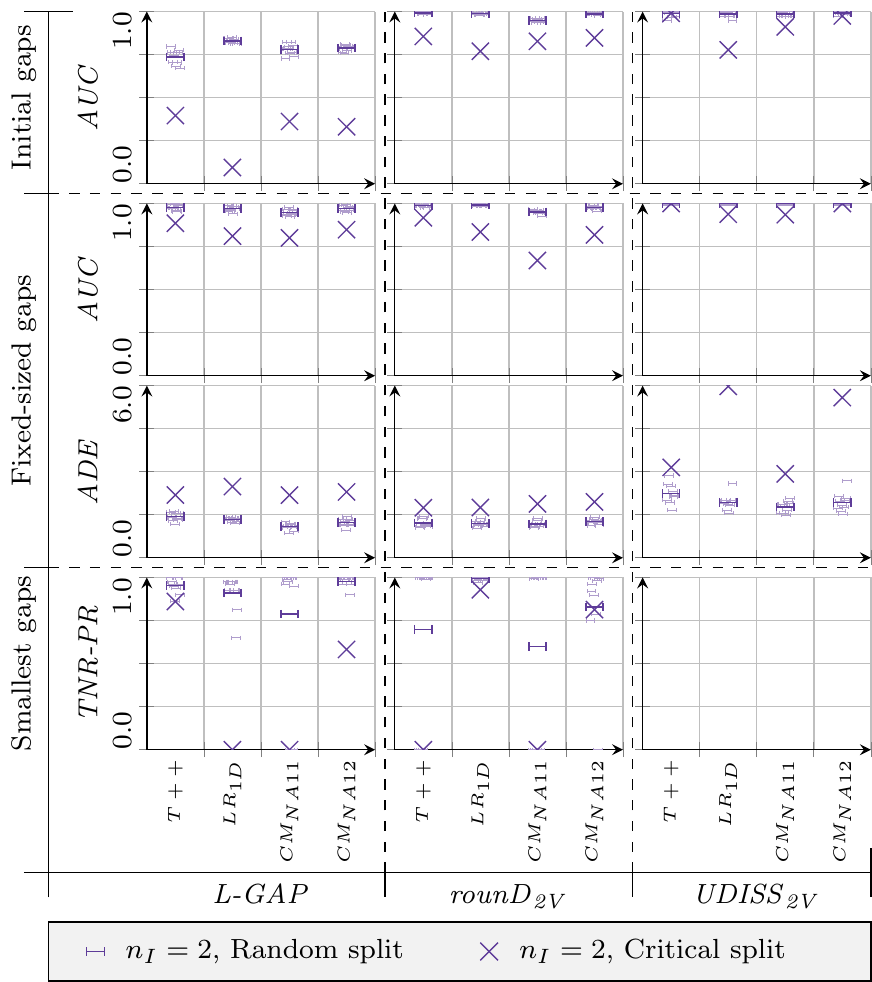}
    \vspace{-5mm}
    \caption{Behavior prediction performance of the two \emph{Commotions} model (\textit{CM}) configurations compared to \textit{Trajectron++} (\textit{T++}) and logistic regression (\textit{LR}) across three datasets (L-GAP, rounD, and Leeds) according to considered metrics (AUC, ADE, TNR-PR). For the random splits, the small markers indicate the results per individual split, while the large markers depict their average.}
    \vspace{-5mm}
    \label{fig:results_III}
\end{figure}
\begin{table}
    \caption{Percentage of cases in which the \emph{Commotions} model performed significantly better or worse compared to the other tested models, based on the results shown in Fig.~\ref{fig:results_III}.  \parnote[]{Notation similar to Tab.~\ref{tab:commmotions_comp}. The configuration $\text{\emph{CM}}_{NA11}$ is used for the \emph{ADE} and $\text{\emph{CM}}_{NA12}$ for the other metrics. The results are split by metric, and for \emph{T++} partially by dataset.}} 
    \begin{tabularx}{\linewidth}{I|X|K|K}
    \toprule
    Model $M$ & \multicolumn{1}{c|}{Cases} & \emph{CM} better than $M$ & \emph{CM} worse than $M$ \\ \midrule
    \multirow{3}{*}{\emph{T++}} & 3 (\emph{ADE}) & $100 \%$ ($0\%$)$\Ho$$\Ho$ & $\Ho \Ho 0\%$ ($0\%$)$\Ho$$\Ho$ \\ 
    & 2 (Other, \emph{UDISS}) & $\Ho \Ho 0\%$ ($0\%$)$\Ho$$\Ho$ &$\Ho \Ho 0\%$ ($0\%$)$\Ho\Ho$ \\ 
    & 6 (Other, Other) & $\Ho 17\%$ ($0\%$)$\Ho$$\Ho$ & $\Ho 33\%$ ($33\%$)$\Ho$ \\ \midrule
    \multirow{2}{*}{$\text{\emph{LR}}_{\text{\emph{1D}}}$} & 3 (\emph{ADE}) & $\Ho 67 \%$ ($67\%$)$\Ho$ & $\Ho \Ho 0\%$ ($0\%$)$\Ho$$\Ho$ \\ 
    & 8 (Other) & $\Ho 13\%$ ($75\%$)$\Ho$ & $\Ho 25\%$ ($0\%$)$\Ho$$\Ho$ \\\bottomrule
    \end{tabularx} 
    \parnotes
    \label{tab:model_comp_1}
    \vspace{-8mm}
\end{table}

Comparison of the models (Fig.~\ref{fig:results_III} and Tab.~\ref{tab:model_comp_1}) demonstrates that the \emph{Commotions} model can compete with established models, although variations were observed depending on the metric and dataset. Notably, the \emph{Commotions} model routinely outperforms the other models in terms of the \emph{ADE}, consistently on the random test sets and, when compared to \emph{LR}, even on the critical test sets. For instance, the average \emph{ADE} achieved by the \emph{Commotions} model on the ten random test sets of the \emph{rounD} dataset is \SI{1.08}{m}, compared to \SI{1.43}{m} for \emph{T++} and \SI{1.33}{m} for \emph{LR}. This may be attributed to the model's capacity to forecast both the probability of accepting a gap and the time at which it may be accepted, with the additional information being used to filter out the most aberrant trajectories suggested by the transformation function (\ref{sec:Framework}).

However, on the other metrics, the \emph{Commotions} model's performance is mostly similar to the other two models (no significant difference on 10/16 random and 8/16 critical splits). Nonetheless, it appears to be more robust than \emph{LR} when predicting unintuitive human behavior, with consistently better outcomes on the critical test. This suggests that constraining a model's predictions using cognitive theory to make it less susceptible to out-of-domain edge cases is a viable way to improve the model's reliability.

The \emph{Commotions} model's worst performance can be observed on the \emph{L-GAP} and \emph{rounD} datasets when compared to \emph{T++} using metrics other than the \emph{ADE}. While this might indicate a superiority of the \emph{T++} model, this deviation in performance may be at least partly explained by the aforementioned differences in the inputs provided to the models. 

To investigate the extent of potential impact of this difference on our results, we compared the second $LR$ model taking two-dimensional inputs to the original $LR$ model processing the one-dimensional inputs. The results of the comparison (Tab.~\ref{tab:expansion_potentials}) show that, at least for the $LR$ model, processing the two-dimensional original inputs (as \emph{T++} does) appears to simplify the prediction task compared to using the quasi-one-dimensional inputs that the \emph{Commotions} model relies on. This seems plausible, as the projection employed to transform the input data from two-dimensional to one-dimensional likely leads to information loss, leaving fewer cues for the models to make accurate predictions. However, more research is required to accurately assess the impact of input dimensionality on predictions. Thus, a final verdict on the comparative advantage of the \emph{Commotions} model or \emph{T++} is still pending.

\section{Conclusion}
This study evaluates the predictive performance of the different configurations of the \emph{Commotions} model, which integrates state-of-the-art theories of human perception, decision-making, and motor control, in gap acceptance scenarios, comparing the best configurations with other established models. The results demonstrate that the \emph{Commotions} model can compete with or even outperform state-of-the-art behavior prediction models, as long as the same input information is provided. Notably, the average displacement error of predicted trajectories is most often significantly lower than the one achieved by other tested models.

We also seek to assess the potential impact of the \emph{Commotions} model's restriction to the quasi-one-dimensional scenario of a perpendicular intersection on its predictive performance. Unable to overcome this restriction, we instead compare two versions of the logistic regression model for this investigation. Our findings suggest that allowing \emph{Commotions} model to instead process two-dimensional trajectories as inputs would be beneficial. As an added benefit, this expansion could also enable the model to function as a dedicated trajectory prediction model. Consequently, such an expansion of the \emph{Commotions} model is likely worthwhile, even if it comes at the cost of more expensive computations. In addition, investigating the impact of other limitations, such as the number of processable input time steps, should be addressed in future research, as it would provide benefits for model designing even beyond the \emph{Commotions} model.

\begin{table}
    \caption{Evaluating the impact of the input dimensionality on the predictive performance of a logistic regression model.\parnote[]{Notation similar to  Tab.~\ref{tab:commmotions_comp}}}
    \begin{tabularx}{\linewidth}{A|A|B|B}
    \toprule
    Dataset & \multicolumn{1}{c|}{Cases} & $LR_{2D}$ better than $LR_{1D}$ & $LR_{1D}$ better than $LR_{2D}$ \\ \midrule
    \emph{UDISS} & 3 & $\Ho 0 \%$ ($33 \%$) & $\Ho 0 \%$ ($0 \%$)$\Ho$ \\
    Other & 8 & $38 \%$ ($13 \%$) & $\Ho 0 \%$ ($13 \%$) \\ \bottomrule
    \end{tabularx} 
    \parnotes
    \label{tab:expansion_potentials}
    \vspace{-8mm}
\end{table}
However, due to its theoretical basis, the \emph{Commotions} model will always be restricted to scenarios such as gap acceptance, where a small number of potential behaviors, like accepting or rejecting a gap, make it feasible to create and evaluate all distinct future trajectories $\widetilde{\bm{\chi}}$. This limits the model's general applicability compared to models like \emph{Trajectron++}. Additionally, as the model itself is non-differentiable, the resulting need for gradient-free optimization makes the model's training process relatively cumbersome, hampering its feasibility further. Nevertheless, our findings provide encouraging evidence supporting the usefulness of the \emph{Commotions} model, at least for predicting human behavior in gap acceptance scenarios, justifying further research into both this specific model and the general approach of integrating cognitive theory into prediction models. For example, it would be worthwhile to investigate how the cognitive assumptions in the \emph{Commotions} model (or other cognitive models) might be leveraged in model architectures that are specifically designed for use in the prediction context.

{\appendices

\input{Appendix}}

\bibliographystyle{jabbrv_ieeetr}
\bibliography{IEEEabrv,main.bbl}

\end{document}

%% file: header.tex
\usepackage[caption=false,font=normalsize,labelfont=sf,textfont=sf]{subfig}
\usepackage{graphicx}
\usepackage{standalone}
\usepackage{tikz}
\usepackage{rotating}
\usetikzlibrary{shapes,backgrounds,shapes.geometric}
\usepackage{circuitikz}
\tikzset{font={\fontsize{9pt}{9}\selectfont}}
\usepackage{pgfplots}
\usepgfplotslibrary{fillbetween,colormaps,groupplots}
\pgfplotsset{compat=newest}

\usepackage[alph, restart, breakwithin]{parnotes}
\usepackage{tabularx}
\usepackage{makecell}
\usepackage{multirow}
\usepackage{booktabs}
\newcolumntype{Y}{>{\raggedleft\arraybackslash}X}
\newcolumntype{Z}{>{\centering\arraybackslash}X}

\newcolumntype{I}{>{\hsize=0.6\hsize}X}
\newcolumntype{J}{>{\hsize=1.2\hsize}X}
\newcolumntype{K}{>{\hsize=1.2\hsize}Z}

\newcolumntype{C}{>{\hsize=1.273\hsize}X}
\newcolumntype{D}{>{\hsize=1.145\hsize}Y}
\newcolumntype{E}{>{\hsize=0.254\hsize}Z}
\newcolumntype{F}{>{\hsize=1.145\hsize}X}
\newcolumntype{G}{>{\hsize=0.636\hsize}X}
\newcolumntype{H}{>{\hsize=1.273\hsize}Z}

\newcolumntype{A}{>{\hsize=0.5\hsize}X}
\newcolumntype{B}{>{\hsize=1.5\hsize}Z}

\newcommand{\Ho}[0]{\hphantom{0}}

\usepackage{siunitx}
\usepackage{amsmath}
\usepackage{amsfonts}
\usepackage{scalerel}
\usepackage{amssymb}
\usepackage{bm}

\usepackage{algorithm}
\usepackage{algorithmicx}

\usepackage{mathtools}
\DeclareMathOperator{\Eqonesep}{\mathrlap{t_i}\hphantom{V_i}}

\usepackage{cite}
\usepackage{jabbrv}

\usepackage[utf8]{inputenc}
\usepackage{enumitem}

\date{\today}

\usepackage[
	pdfauthor={Julian Schumann},
	pdfcreator={pdfLatex},
	pdfsubject={First project for PhD},
    colorlinks={true},
    linkcolor={blue},
    citecolor={blue},
    urlcolor={red},
    hyperindex={true}
]{hyperref}
\makeatletter
\hypersetup{pdftitle=\@title}
\makeatother
